\documentclass[letterpaper, 10 pt, conference]{ieeeconf}  

\IEEEoverridecommandlockouts                              

\usepackage{xstring}
\usepackage{graphicx}
\usepackage{pgfplots}
\usepackage{pgfplotstable}
\usepackage{tikz}
\usepackage{subfig}
\usepackage{booktabs}
\usepackage{multirow}
\usepackage{url}
\usepackage{amsmath}
\usepackage{amssymb}
\usepackage{pifont}

\newcommand{\cmark}{\ding{51}}\newcommand{\xmark}{\ding{55}}

\newcommand{\real}{\mathbb{R}}

\newcommand{\SE}[1]{\mathrm{SE}(#1)}

\renewcommand{\vec}[1]{\mathbf{#1}}

\newcommand{\norm}[1]{\lVert #1 \rVert}

  \newcommand{\inverse}[1]{{#1}^{-1}}
\newcommand{\transpose}[1]{{#1}^{\mathsf T}}

\newcommand{\intrinsics}{\mathsf{K}}

\newcommand{\pose}{\mathcal{P}}

\newcommand{\keyframe}{\mathcal{K}}

\newcommand{\refFigure}[1]{Fig. \ref{#1}}

\newcommand{\refSection}[1]{section \ref{#1}}

\newcommand{\refTable}[1]{Table \ref{#1}}

\newcommand{\etAl}{\emph{et al.}\mbox{ }}

\title{\LARGE \bf
Semantic Image Alignment for Vehicle Localization
}

\author{Markus Herb$^{1,2,\ast}$, Matthias Lemberger$^{2,1,\ast}$, Marcel M. Schmitt$^{2}$, Alexander Kurz$^{2}$,\\ Tobias Weiherer$^{2}$, Nassir Navab$^{1}$ and Federico Tombari$^{1,3}$\thanks{$^{1}$Technical University of Munich, Department of Informatics, Germany}\thanks{$^{2}$AUDI AG, Ingolstadt, Germany}
\thanks{$^{3}$Google, Zurich, Switzerland}\thanks{$^{\ast}$Equal Contribution}}

\begin{document}
\maketitle%
\thispagestyle{empty}%
\pagestyle{empty}%

\tikz[remember picture,overlay] {%
    \node at (current page.south) {	
    	\raisebox{2cm}{
            \parbox{\textwidth}{
            \footnotesize
            \centering
            \copyright~2021 IEEE. Personal use of this material is permitted. Permission from IEEE must be obtained for all other uses, in any current or future media, including reprinting/republishing this material for advertising or promotional purposes, creating new collective works, for resale or redistribution to servers or lists, or reuse of any copyrighted component of this work in other works.
}}
    };%
    }%
\begin{abstract}%
Accurate and reliable localization is a fundamental requirement for autonomous vehicles to use map information in higher-level tasks such as navigation or planning.
In this paper, we present a novel approach to vehicle localization in dense semantic maps, including vectorized high-definition maps or 3D meshes,
using semantic segmentation from a monocular camera.
We formulate the localization task as a direct image alignment problem on semantic images, which allows our approach to robustly track the vehicle pose in semantically labeled maps by aligning virtual camera views rendered from the map to sequences of semantically segmented camera images.
In contrast to existing visual localization approaches, the system does not require additional keypoint features, handcrafted localization landmark extractors or expensive LiDAR sensors.
We demonstrate the wide applicability of our method on a diverse set of semantic mesh maps generated from stereo or LiDAR as well as manually annotated HD maps and show that it achieves reliable and accurate localization in real-time. 
\end{abstract}

\section{Introduction}

Highly detailed maps of the road infrastructure are considered a crucial enabler for autonomous vehicles and most autonomous driving systems make extensive use of maps to aid navigation and action planning.
Detailed semantic maps offer a dramatically improved understanding of the vehicle environment by augmenting the available onboard sensor data with information about road area, traffic signs, road markings, obstacles and more.
This is particularly valuable in challenging situations such as dense urban traffic, heavily occluded scenes as well as at large distances, where sensor performance typically degrades.
Commercial mapping providers are developing increasingly detailed 3D high-definition maps for autonomous vehicles with rich semantic information that are highly abstracted from underlying sensor data, which allow for lightweight storage, easy interpretation and sharing across different vendors.

To exploit the map knowledge, a precise localization within the map is of tremendous importance and should work reliably in any condition including GPS-denied areas or difficult urban regions.
Localization is a fundamental problem in robotics and has been addressed with visual localization methods in the past, which typically rely on specific handcrafted or learned features as localization landmarks.
In many real-world applications including autonomous vehicles however, the maps do not contain landmarks designed for the task of localization, because they have been created with different methods or sensors, or are provided by third parties.

\begin{figure}[tp]
    \centering
    \includegraphics[width=0.49\columnwidth]{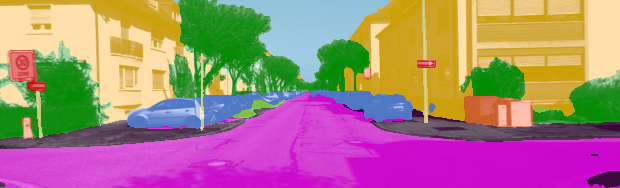}~
    \includegraphics[width=0.49\columnwidth]{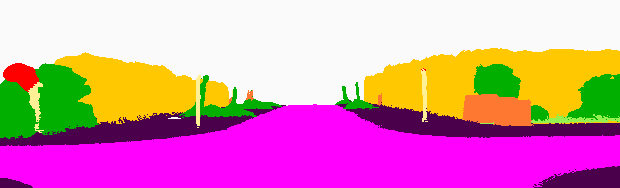}
    \vspace*{-0.25cm}
    \\
    \includegraphics[width=1.0\columnwidth]{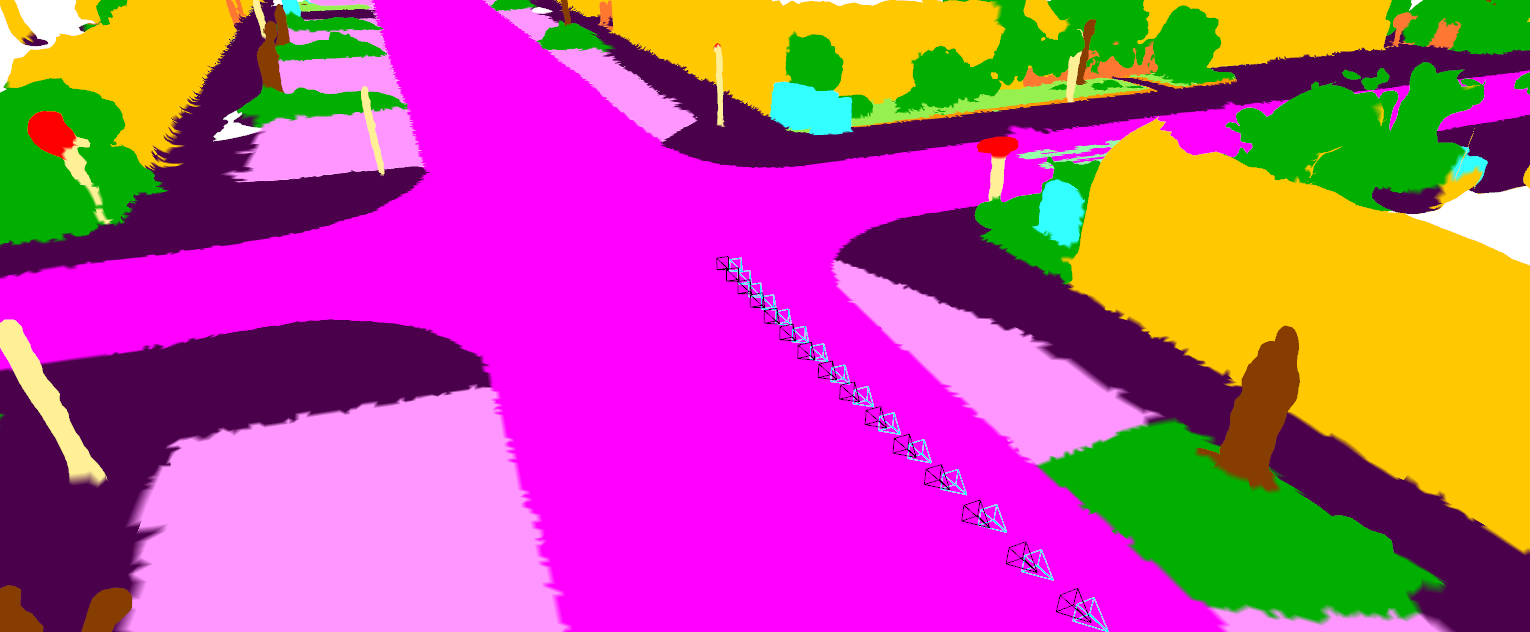}
    \caption{Camera-based vehicle localization in a dense semantic map (bottom, estimate in cyan, ground truth in black) by aligning semantically segmented camera frames (top left) to rendered virtual views of the map (top right)}
    \label{fig:teaser}
 \end{figure}
 
A promising way to overcome the lack of specific localization landmarks is to directly use the semantic and geometric information, which are provided for higher-level tasks, also for localization purposes.
In addition to enable localization in various existing and future map formats, using semantic information as localization landmarks yields many other appealing properties.
Semantic information can easily be obtained from sensor data, since segmentation networks are available for different sensor modalities and are often readily available in the system pipeline as they are used for other perception tasks.
Furthermore, semantics is largely invariant to different environmental conditions, such as weather or season, and across different sensor modalities.

Our localization approach is inspired by the success of direct image alignment methods for odometry estimation \cite{Steinbrucker2011} and map-relative tracking \cite{Newcombe2011}.
The fundamental idea driving our method is to recover a precise vehicle pose with respect to the map by aligning camera frames to virtual map views rendered from the semantic map.
Starting from an initial pose estimate obtained from e.g. GPS or place recognition, our system directly optimizes for the vehicle pose, which enables localization in real-time.
Our approach offers high practical value, as it only requires semantically segmented camera images and ubiquitous odometry input for localization and does not need special landmarks for localization.
This makes our system appealing for a wide range of application areas with a diverse set of map formats from indoor augmented reality applications to autonomous vehicle navigation in large-scale outdoor maps.

Our main contributions in this work are threefold:

\begin{itemize}
    \item A camera-based localization approach using image-alignment exploiting semantic map elements as landmarks
    \item A novel probabilistically sound semantic image alignment loss to maximize consistency between the map and sensor perception
    \item An extensive evaluation on three datasets with diverse map formats and ablation studies to show the efficacy of our localization method
\end{itemize}

\section{Related Work}
In the following, we present prior work related to our approach. Considering the large amount of work in the field of vehicle localization, we focus our review on image-based methods as well as relevant approaches using either semantics or direct image alignment.

\subsection{Vehicle Localization}

Sparse keypoint features based on feature detectors and descriptors are a well researched method for map reconstruction and localization in Visual SLAM and have also been proposed for localization of automated vehicles \cite{Lategahn2014}.
Improvements for robustness have been proposed recently \cite{DeTone2018, Sarlin2020}, but a main limitation remains that dedicated localization features are required and that the map must be created from camera images.

As an alternative to using image-based localization features, naturally occurring elements in road environments such as lane markings \cite{Schreiber2013}, pole-like objects \cite{Spangenberg2016}, traffic signs or combinations of such features \cite{Qu2018, Kummerle2019, Ma2019} have been proposed as localization landmarks.
All of these methods require detectors engineered or trained to detect the specific features and may get lost easily in areas where few of such landmarks are present, for instance in rural or residential areas.
Our method instead relies on common and ubiquitous semantic segmentation classes to perform a matching to the map.

Another line of localization approaches rely on dense point clouds as maps and perform localization using geometric alignment \cite{Caselitz2016}.
CMRnet \cite{Cattaneo2019} instead learns to directly match image points to projected 3D point clouds to perform localization using 2D-3D pose estimation.

\subsection{Semantic Localization}

With the tremendous improvements of semantic image segmentation driven by deep learning in recent years, several works leveraged semantics for place-recognition and metric localization.
VLASE \cite{Yu2018a} is an image-based place recognition system that uses the edges between different semantic regions as robust features for image retrieval.
Toft \etAl \cite{Toft2017} demonstrated single-image metric localization in a 3D model of semantic points and curves, by minimizing the distance of projected map elements to a matching semantic segment in image space.
\cite{Pauls2020} extended this idea to map-based localization over multiple frames.
While these methods ground on similar ideas as our method, we leverage image alignment to achieve high efficiency, accuracy and robustness through dense alignment and multi-scale optimization.

Several methods were proposed in \cite{Toft2018, Shi2019, Garg2019} that enhanced traditional keypoint-based localization in a 3D model by using semantics to improve image-retrieval or by using consistency checking for more reliable 3D pose-estimation.
However, these still require image keypoint features as additional localization cues and cannot work directly with purely semantic maps.
Sch\"onberger \etAl \cite{Schoenberger2018} learned descriptors for locally aggregated semantic 3D maps for robust long-term place recognition and used further 3D alignment for metric localization, but requires depth for aggregating local maps and focuses more on global relocalization.
In \cite{Stenborg2018}, a particle filter-based localization system was proposed that uses a sparse 3D map of semantically labeled points for localization by evaluating the consistency of semantic points projected to image space for each particle.
We follow the same idea of using temporal information to improve the robustness of semantic localization, but instead of using particle filtering, we formulate the problem in a computationally more efficient non-linear least squares framework.

\subsection{Direct Image Alignment for SLAM}

Our localization approach presented in this work is strongly inspired by recent work in the field of Direct Image Alignment for visual odometry and SLAM tasks.
In Direct Image Alignment methods, the relative camera pose between two frames is found by warping the intensity image from one frame to the other and minimizing the resulting photometric error assuming that the brightness in the scene is constant.

Comport \etAl \cite{Comport2007} proposed a stereo-based visual odometry system exploiting this concept, which was also used for visual odometry \cite{Steinbrucker2011} and SLAM \cite{Kerl2013} using RGB-D sensors.
Image Alignment methods were also proposed for monocular setups, with DTAM \cite{Newcombe2011} able to reconstruct a dense map and tracking the camera motion against rendered views of this map.
Engel \etAl later presented efficient monocular-based SLAM \cite{Engel2014} and odometry \cite{Engel2018} methods that simultaneously estimate a semi-dense or sparse depth map and the camera pose for each keyframe.

\begin{figure*}[t!]
\centering
\resizebox{\textwidth}{!}{
\includegraphics{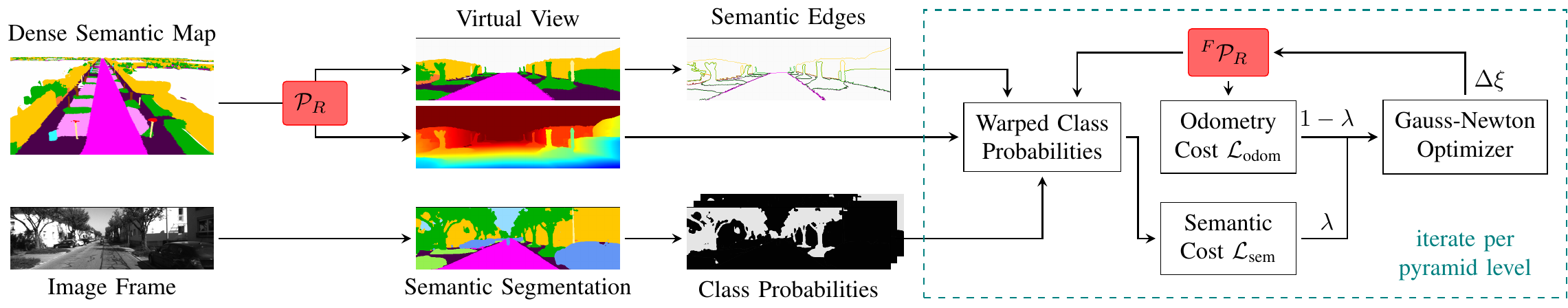}
 }
    \caption{Overview of the proposed Semantic Image Alignment pipeline. A virtual map view with semantics and depth is rendered at pose $\pose_R$, from which semantic edge pixels are extracted. The image frame is semantically segmented and converted to class probabilities. Relative pose $^F\pose_R$ is updated iteratively using pose increments $\Delta\xi$ computed from warped class probability residual loss measuring semantic consistency and additional odometry constraints. Actions in the dashed box are repeated for each iteration and scale. Multi-scale and keyframe windowing omitted for clarity.}
    \label{fig:pipeline}
\end{figure*}

While photometric alignment of images has shown outstanding results for tasks such as visual odometry, the largest drawback of these methods is their susceptibility to lighting changes, which makes medium to long-term relocalization challenging.
In \cite{VonStumberg2019a}, the authors propose to use deep-learned feature representations instead of directly using image intensities to robustify the image alignment for long-term relocalization.
Similar to feature-based approaches, this approach however comes with the limitation that dedicated high-dimensional localization descriptors need to be trained, computed and stored in the map as an additional layer.
In our approach instead, we rely on commonly available semantic information as robust localization features, which is oftentimes readily available both in existing maps and online due to its use in other perception components.
While other works \cite{Mahe2018a} have used semantics in image-alignment based systems to robustify RGB-D based visual odometry, the method presented in this paper is the first method that uses direct image alignment with semantic segmentation for map-based localization.
 \section{Semantic Image Alignment}

The task of a localization system in automated vehicles is to recover the precise vehicle position within an existing 3D map such that higher level modules can use the available map data.
To this end, we require the algorithm to output a metric 6 DoF pose $\pose_F \in \SE{3}$ for each frame $F$.

An overview of our localization algorithm is depicted in \refFigure{fig:pipeline}.
The fundamental idea behind our localization system is to estimate the relative pose $^F\pose_R$ between a perceived camera frame $F$ and a \emph{virtual} camera view $R$ rendered from the 3D map at a known initial estimate of the vehicle pose $\pose_R$.
Given this, we obtain the camera pose in the map as
\begin{equation}
    \pose_F = \pose_R\cdot\inverse{^F\pose_R}.
\end{equation}
We find this relative pose $^F\pose_R$ iteratively by warping the semantic information in frame $F$ into the rendering $R$ using the depth information from the rendering and optimize for the warped semantics to be consistent with the rendered virtual view.
Using the rendered depth for warping is convenient, because we obtain accurate depth information and do not require any online depth computation.

Starting from an initial pose estimate, we continuously track the vehicle position within the map at defined keyframes for each of which we render a virtual view based on the pose of the previously aligned keyframe.
To keep the pose tracking temporally consistent, we stabilize the pose estimation by enforcing odometry constraints between consecutive keyframes.

\subsection{Direct Image Alignment}
Since our approach is based on the concept of Direct Image Alignment \cite{LukasKanade1981}, we begin with a brief introduction of the general concept, before presenting our semantic image alignment in \refSection{sec:semantic_alignment_loss}.
Direct Image Alignment has shown great success mostly in RGB-D based odometry \cite{Steinbrucker2011} and SLAM \cite{Kerl2013} among other applications and is, in a SLAM context, understood as the task of finding the relative transformation $^2\pose_1 \in \SE{3}$ between two camera frames $F_1$ and $F_2$ that aligns the images of both frames with minimal error.
We assume two camera frames $(\mathit{I}_1, \mathit{D}_1)$ and $(\mathit{I}_2, \mathit{D}_2)$ with intensity image $\mathit{I} \in \real^{H\times W}$ and corresponding depth image $\mathit{D} \in \real^{H\times W}$.
Further we use the warp function $^2w_1(^2\pose_1, \vec{x}) : \SE{3} \times \real^2 \to \real^2$ that computes pixel locations in frame $F_2$ given source pixel location $\vec x$ and corresponding depth $\mathit{D}_1(\vec x)$ in frame $F_1$ and relative transform $^2\pose_1 \in SE(3)$: 
\begin{align}
    \label{eq:warp}
^2w_1(^2\pose_1, \vec{x}) &= \pi_\intrinsics\left(^2\pose_1 \cdot \inverse{\pi_\intrinsics}\left( \vec x, \mathit{D}_1(\vec x) \right)\right)
\end{align}
with the unprojection function
\begin{equation}
    \label{eq:unproject}
    \vec{X}_1 = \inverse{\pi_\intrinsics}(\vec x, d) = d \cdot \inverse{\intrinsics} \transpose{[\vec x, 1]}
\end{equation}
given camera intrinsics $\intrinsics$.
The projection function $\pi_\intrinsics$ is defined as
\begin{equation}
    \label{eq:project}
    \vec{x}_2 = \pi_\intrinsics(\vec{X}_2)
    = \mathrm{pr}\left(\intrinsics \cdot \vec{X}_2\right)
\end{equation}
with $\mathrm{pr}\left({\transpose{[x, y, z]}}\right) := \transpose{[x/z, y/z]}$.

Using the warp function, we define the squared image alignment loss
\begin{equation}
    \sum\limits_{i} \norm{ I_2(^2w_1(^2\pose_1, x_i)) - I_1(x_i) }^2 = \sum\limits_{i} \transpose{r_i}r_i
\end{equation}
over pixels $i$.
We choose to parametrize the relative transform as twist coordinates $^2\xi_1 \in \real^6$ for a minimal number of 6 parameters with $^2\pose_1 = \exp(^2\xi_1)$ using the exponential map $\exp : \real^6 \to \SE{3}$ and obtain the final residual
\begin{equation}
    \label{eq:image_align_residual}
    r_i(^2\xi_1) = I_2(^2w_1(\exp(^2\xi_1), x_i)) - I_1(x_i).
\end{equation}
A solution to the above non-linear least squares problem is typically found using Gauss-Newton optimization by iterating
\begin{equation}
    \xi^{(k+1)} = \xi^{(k)} - \Delta\xi
\end{equation} until a sufficient solution has been reached.
The update $\Delta\xi$ in each iteration $k$ is computed by finding a solution to 
\begin{equation}
    \transpose{J_r}J_r \cdot \Delta\xi = \transpose{J_r}\vec r
\end{equation}
with $J_r$ being the $N \times 6$ stacked Jacobian of residual derivatives $\frac{\partial}{\partial \xi} r_i(\xi^{(k)})$ and $\vec{r}(\xi^{(k)})$ being the $N \times 1$ vector of residuals, each evaluated at the current estimate $\xi^{(k)}$. We compute all jacobians analytically using the chain rule.

\subsection{Semantic Alignment Loss}
\label{sec:semantic_alignment_loss}
After introducing the general concept of image alignment in SLAM, we will now present our novel semantic image alignment loss.
Instead of optimizing for photometric consistency, we aim to optimize for semantic consistency of the aligned frames.
Semantic consistency is typically, e.g. in training a semantic segmentation neural network, enforced using the cross-entropy loss, which maximizes the softmax-probability
\begin{equation}
    \sigma_c(x) = \frac{\exp(x[c])}{\sum_{i=0}^{n-1} \exp(x[i])}
\end{equation}
for the true class label $c$ given logit predictions $x \in \real^N$ for each of the $N$ classes.
In our semantic alignment, we determine the semantic consistency by warping the class logits from the frame semantics $F$ into our rendering $R$ and evaluate the softmax-probability for the class label in the rendered frame.

In order to re-use the methods developed for image alignment within our semantic localization, we formulate the semantic consistency optimization in a least-squares framework.
We start off with the probability density $p(x, c)$ given by the softmax $\sigma$ over the logits $x$ for a given class $c$ and re-write as
\begin{align}
    p(x, c) &= \sigma_c(x) = \exp(\log(\sigma_c(x))) \nonumber \\
    &= \exp\left(-\frac{1}{2}(-2\log(\sigma_c(x)))\right) \nonumber \\
    &= \exp\left(-\frac{1}{2}\sqrt{-2\log(\sigma_c(x))}^2\right)
\end{align}
We note that this density is directly proportional to a standard normal distribution $\mathcal{N}(t; 0,1)$ evaluated at $t = \sqrt{-2\log(\sigma_c(x))}$ and as such we obtain a maximum likelihood solution for $p(x, c)$ in a least-squares framework by using $r(x, c) = \sqrt{-2\log(\sigma_c(x))}$ as our residual.
Putting this into our image-alignment problem, we optimize for the semantic consistency loss
\begin{equation}
    \mathcal{L}_\text{sem}(^F\xi_R, \vec{x}_i, c_i) = 
    -2\log(\sigma_{c_i}(w(\exp(^F\xi_R), \vec{x}_i)))
\end{equation}
over pixels $i$ with coordinates $\vec{x}_i$ and class label $c_i$ in the rendered frame $R$.

\subsubsection{Multiscale Optimization}
As customary in image alignment problems, we optimize over multiple image scales to avoid local minima in optimization space and improve convergence.
We build an image pyramid with scaling factor 2 and optimize iteratively starting from the lowest resolution over the lowest few levels, because we found higher scales did not improve the results significantly.
We re-scale the virtual semantic and depth views using nearest neighbor interpolation.
For the class probabilities from the semantically segmented image, we first compute a logits image $L \in \real^{H\times W \times N}$ for $N$ classes.
For each level, we re-scale the logits image to $\real^{H/2\times W/2 \times N}$ by computing the mean logits value in each $2 \times 2$ pixel cell for each class and then apply the softmax, which increases the convergence radius compared to nearest neighbor scaling.

\subsubsection{Edge Pixel Selection}
Computing the residual for every pixel in an image incurs a significant runtime cost.
Similar to existing alignment based methods such as DSO \cite{Engel2018}, we select only relevant pixels with a high gradient in image space for computing the residuals.
For our semantic alignment, we choose every pixel adjacent to an edge between two different semantic regions in the rendered image, excluding pixels labeled as background.
Such edge pixels discriminate between different semantic regions in the scene and typically fall into high-gradient regions in the corresponding class-probability image of the camera frame, leading to fast and accurate convergence.

\subsection{Windowed Optimization}

The accuracy of aligning a single frame to a rendered view dramatically depends on the semantic complexity or richness of a scene.
For scenes with very little semantic content or large occlusions, the semantic borders in a single view may not be enough to align the camera frame to the map with sufficient accuracy.
Similar to other odometry and localization approaches, we therefore choose to perform the alignment within an active window $\mathcal{W}$ of keyframes $\left\{\keyframe_1, \ldots, \keyframe_n\right\}$ to keep temporal consistency and hence stabilize the pose estimates.
To keep the method generic, we rely on external odometry input for each frame, which can be supplied e.g. by inertial sensors, a vehicle odometry subsystem or a visual odometry method.

For each pair of consecutive keyframes $(\keyframe_k, \keyframe_{k+1})$ in the window $\mathcal{W}$ we formulate an odometry cost as 
\begin{equation}
    \mathcal{L}_\text{odom}(\xi^{(k)}, \xi^{(k+1)}) = 
    \left\lVert
    W_O
    \left(
    ^{(k)}\pose_{(k+1)}^O
    \boxminus
    ^{(k)}\pose_{(k+1)}^E
    \right)
    \right\rVert^2
\end{equation}
using $\mathcal{A} \boxminus \mathcal{B} := \log(\inverse{\mathcal{B}}\mathcal{A})$ that penalizes deviations of the estimated relative motion after the optimization given as $^{(k)}\pose_{(k+1)}^E$ from the measured odometry $^{(k)}\pose_{(k+1)}^O$.
The residual is weighted with a diagonal weight matrix $W_O$ modeling the odometry noise.
The optimized relative motion is computed as
\begin{equation}
    ^{(k)}\pose_{(k+1)}^E = \inverse{\left(\pose^{(k)}_F\right)} \cdot \pose^{(k+1)}_F
\end{equation}
with $\pose^{(k)}_F = \pose^{(k)}_R \cdot \inverse{\exp(^F\xi_R^{(k)})}$.

Combining the semantic and odometry costs in the window, we obtain the total cost function
\begin{align}
    \mathcal{L}_\text{loc}\left(\vec{\xi}\right) =&
    \lambda \sum\limits_{k=1}^{n}\sum\limits_{i} \mathcal{L}^{(k)}_\text{sem}\left(\xi^{(k)}, \vec{x}_i^{(k)}, c_i^{(k)}\right)
    \nonumber \\
    &+
    (1-\lambda)\sum\limits_{k=1}^{n-1} \mathcal{L}^{(k)}_\text{odom}\left(\xi^{(k)}, \xi^{(k+1)}\right)
\end{align}
over all active keyframes $\keyframe_k \in \mathcal{W}$ and rendered edge pixels $i$ in each $\keyframe_k$ with weighting factor $\lambda \in [0, 1]$.
For the optimization window, we set a new keyframe every 200ms, shift the optimization window and start the optimization in the background.
Poses of intermediate incoming frames are extrapolated w.r.t. the last optimized keyframe in the window based on odometry input.
 \section{Evaluation}

\newcommand{\plotcdfplot}[3]
{
   \addplot[index of colormap={\namedcolor{#2} of \cmap}, #3] table [x=x, y=y] {#1.dat};
}

\subsection{Implementation}
We implemented the proposed system in Python with some performance-critical parts optimized using C++ bindings and virtual map view rendering through OpenGL bindings.
We employ PSPnet \cite{Zhao2017} to compute the image frame segmentations, which we train and run on a Tesla V100 GPU.
Instead of using the raw logits predictions from PSPnet, we convert per-pixel class predictions to logits by assigning the predicted class a softmax probability of 0.9 and equal probability to all remaining classes.
For all datasets presented in the following, we chose to use 6 image pyramid levels and compute the alignment on the coarsest 3 levels with 10 iterations on each level, because we found including finer levels did not improve the results significantly.
The optimization window includes 8 keyframes.
We benchmarked the algorithm on an Intel Core i7-6820HQ laptop CPU and measure an average runtime of around 180ms for each keyframe optimization using only a single core, with negligible runtime for each extrapolated intermediate frame.
Given a keyframe-rate of 5 keyframes per second typical for visual odometry systems, our algorithm hence runs in real-time.

\subsection{Datasets}
\label{sec:eval_datasets}
We evaluate our system on three different datasets to show the versatility of our proposed approach:
\paragraph{KITTI} The KITTI dataset \cite{Geiger2012} is a well-known AD dataset, which we use to demonstrate camera-based localization in a map generated using LiDAR data, i.e. localization across different sensor modalities.
Since each route was recorded only once, we follow the evaluation procedure of \cite{Caselitz2016} \cite{Cattaneo2019} and consider camera and LiDAR modalities as independent by using only LiDAR data for map creation and only camera data for localization.
We generate a map by aggregating a semantically labeled LiDAR point cloud using ground truth labels from SemanticKITTI \cite{Behley2019}, which is then converted to a semantic mesh using the Poisson Surface Reconstruction in Colmap \cite{Schonberger2016a}.
For camera semantics, we pre-trained our segmentation network in CityScapes \cite{Cordts2016} and fine-tuned on the official KITTI semantic segmentation labels.
Raw velocity and angular rate estimates are used as odometry input with $\lambda=0.65$.
\paragraph{Oxford RobotCar} The Oxford RobotCar dataset \cite{Maddern2017} consists of sequences of the same route collected under different weather conditions, of which we use the benchmark sequences proposed in \cite{VonStumberg2019a} for long-term localization.
We pre-trained a camera segmentation network on A2D2 \cite{aev2019} and fine-tuned on 90 hand-labeled frames selected from Oxford sequences not contained in the benchmark to bridge the domain gap.
We use the semantics component of Kimera \cite{Rosinol2019a} to compute a dense semantic TSDF reconstruction with 10cm voxel size using stereo-depth and camera semantics predicted by PSPnet, from which we extract the final semantic mesh as map. As odometry input to our system, we use the stereo visual odometry results in the dataset and $\lambda=0.8$.
\paragraph{High-Definition Maps} We further collected an additional dataset based on a commercially available high-definition map generated from hand-labeled and geo-referenced LiDAR and camera data. We recorded sequences of around 10km total length using an automotive-grade front-facing grayscale camera, vehicle odometry and RTK-GPS as ground truth. $\lambda=0.5$ is used as parameter. We use this dataset to show the performance in high-definition maps with a high level of sensor abstraction.

\subsection{Baselines}
We compare our method against several different, state-of-the-art methods for localization covering a broad range of different approaches.

\paragraph{Semantic Particle Filter}
We compare against existing semantic localization methods using a Semantic Particle Filter, similar to the one proposed in \cite{Stenborg2018}.
Instead of using a sparse set of semantic 3D points, we re-use the same dense 3D map reconstruction as for our approach and again render a virtual semantic view of the scene for each particle.
We score each particle by comparing the rendered view to the actual frame segmentation.
In all experiments, we use 500 particles for runtime reasons and choose the mean of the best 10\% of particles as the localization result.
As in our approach, we choose every fifth frame as a keyframe and extrapolate intermediate poses.

\paragraph{Feature-based Localization}
To compare our approach to feature-based localization, we use hloc \cite{Sarlin2018a} with SuperPoint \cite{DeTone2018} and SuperGlue \cite{Sarlin2020} as a strong state-of-the-art baseline.
We triangulate a 3D model from the 20 closest neighbor frames including left and right stereo image given camera poses from visual odometry.
In addition, we also list the re-localization results of ORB-SLAM \cite{Mur-Artal2017a}.

\paragraph{Direct Image Alignment}
We further include the results of Direct Sparse Odometry \cite{Engel2018} as well as GN-Net \cite{VonStumberg2019a} from the GN-Net Benchmark. While DSO represents a traditional, image-based direct image alignment method, GN-Net uses improved deep-learned invariant features for long-term localization.

\paragraph{Lidar-Camera Localization}
For our LiDAR-based map in KITTI, we include CMRNet \cite{Cattaneo2019} as a baseline, which localizes images against pre-accumulated LiDAR point cloud maps using learned point-matching. We use the pre-trained model provided by the authors.

\subsection{Results}
In the following, we discuss the quantitative relocalization results for the different datasets and comparisons to baseline methods. For qualitative results, we kindly refer to our supplementary video.

\subsubsection{KITTI}
\begin{figure}[thpb]
   \centering
   \includegraphics{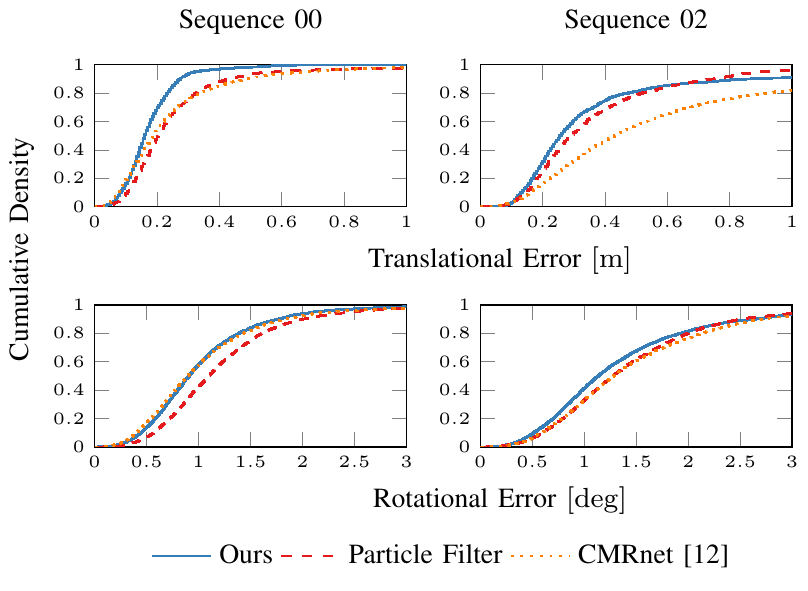}
\caption{
   Cumulative error distribution on KITTI sequences 00, 02 for the translation (top) and rotation (bottom).
}
   \label{fig:eval_kitti_cum}
\end{figure} 

In \refFigure{fig:eval_kitti_cum}, we show the absolute translational and rotational errors for two KITTI sequences.
Since the maps are based on LiDAR data, we list CMRNet \cite{Cattaneo2019} and the particle filter implementation as baselines.
We can see that our proposed method is on par or outperforming the baselines in each sequence for both translation and rotation. By consistently reporting a median relocalization error of around 20cm across all sequences, it demonstrates its suitability for most automated driving tasks.

\subsubsection{Oxford RobotCar}
\begin{figure}[thpb]
   \centering
   \includegraphics{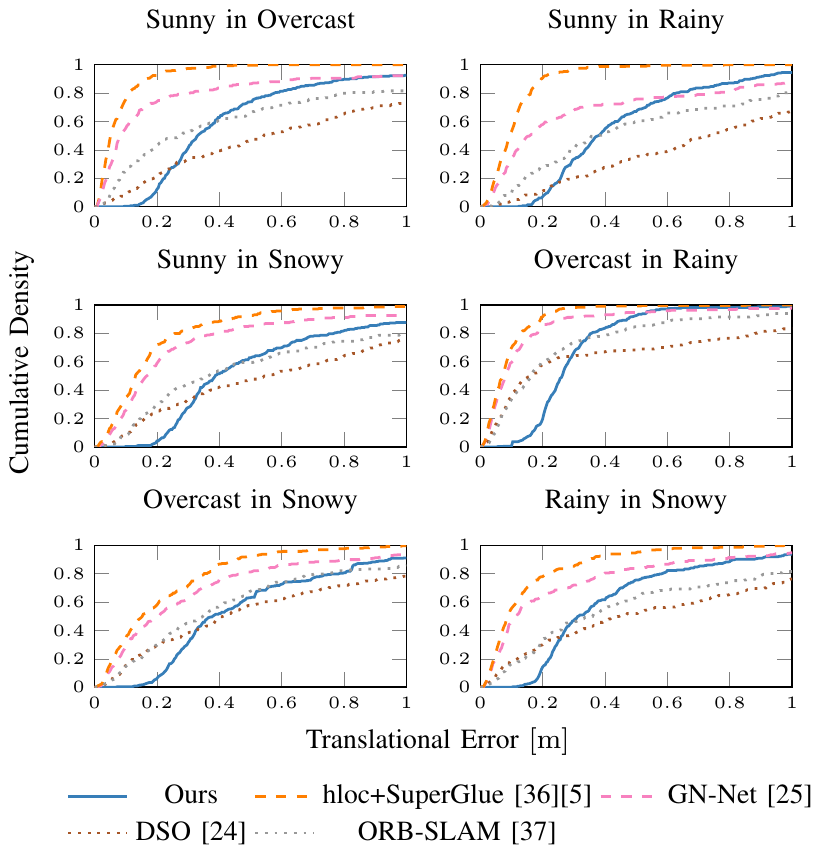}
\caption{Cumulative localization errors on GN-Net Benchmark sequences of Oxford RobotCar. We compare our approach against
   state-of-the-art feature-based (hloc, ORB-SLAM) and direct (GN-Net, DSO) visual localization methods.}
   \label{fig:eval_oxford_cum}
\end{figure}
 For each of the GN-Net Benchmark \cite{VonStumberg2019a} sequences of the Oxford RobotCar dataset, the relocalization errors are depicted in \refFigure{fig:eval_oxford_cum}.
Comparing the results against a number of different image-based localization methods, we observe further that our method shows competitive to superior performance with respect to traditional direct or feature-based localization methods, while the proposed system is unable to reach the performance of recently proposed optimized methods based on learned image feature descriptors such as hloc with SuperGlue or GN-Net. 
This is mainly due to the fact that semantics as intermediate feature representation is, as one would expect, inferior to feature representations specifically trained for the task.
However, semantics as used in our approach on the other hand comes with the advantage of potentially much reduced storage requirements, invariance across sensor modalities and map formats and can further be used for other perception tasks within autonomous driving pipelines.
We provide an additional ablation study with respect to map and semantics quality on this dataset in \refSection{sec:eval_ablation_oxford}.

\subsubsection{HD-Map}
\begin{figure}[thpb]
   \centering
   \includegraphics{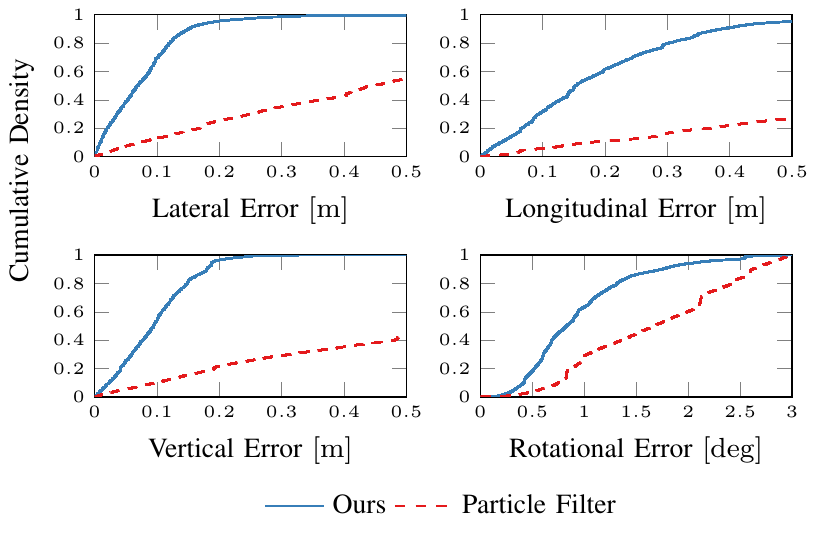}
\caption{Cumulative localization errors in High-Definition Map sequences, split into lateral (top left), longitudinal (top right) and vertical (bottom left) vehicle direction as well as orientation (bottom right).}
   \label{fig:eval_d5_cum}
\end{figure}
 We list the localization results on our High-Definition Map sequences in \refFigure{fig:eval_d5_cum}, split by the individual degrees of freedom.
First, we observe dramatic improvements compared to the particle filter baseline for all individual error components and much more pronounced quantitative difference than observable in the KITTI benchmark.
We believe this is due to the fact that the HD Map contains a less dense environment representation, which is because the data are hand-labeled for the most significant elements of the environment and due to strong compression into condensed vectorized maps.
While the particle filters seems to struggle to cope with these more challenging conditions, our approach does not degrade in performance, because we explicitly focus on the boundaries between different semantic regions to estimate the vehicle pose.

Further we can see in this experiment that the lateral and vertical error components are subject to a much lower error compared to the longitudinal component.
While the vertical component is typically of less importance in automated driving, lateral and longitudinal errors are of great importance for the dynamic driving task.
With a reported lateral error of less than 10cm close to 80\% of the time and always less than 25cm as well as a longitudinal error of less than 50cm almost throughout the entire journey, the localization accuracy is well within the needs for urban automated driving. 

\subsubsection{Initialization Behavior}
\begin{figure}[thpb]
   \centering
   \includegraphics{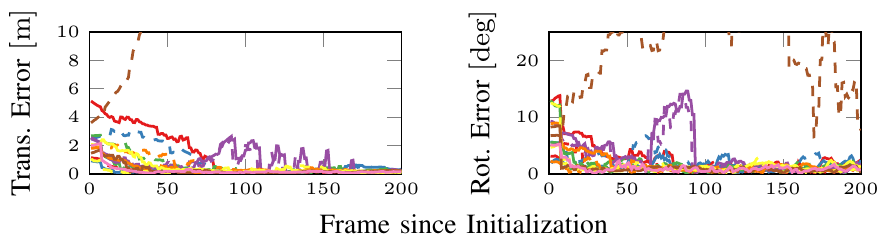}
\caption{Time-series plot of translational (top) and rotational (bottom) errors for 15 randomly initialized starting poses in KITTI sequence 00 with offsets up to 5 meters and 15 degrees from the true position. With the exception of one initialization, all initializations converge to a proper solution.}
   \label{fig:eval_kitti_init}
\end{figure}
 A common problem in methods using image alignment is the fact that the optimization procedure is very sensitive to the initial relative pose estimate.
Particularly for map-based relocalization, this can become an issue when the initial pose supplied by a GNSS or place recognition is not very reliable.
In order to evaluate the behavior of our system in case of noisy initialization, we observe the localization errors after 15 randomly initialized poses within KITTI sequence 00 with random offsets up to 5m and up to 15deg from the ground truth pose.
The error time series are depicted in \refFigure{fig:eval_kitti_init}.
We observe that, while a single initialization diverges entirely, all others converge to a proper solution with small errors after at most several dozens of frames.
This shows that our system can properly track the vehicle location in most cases, even when initialized with multiple meters or degrees of initial offset.

 \subsection{Ablation Study}
\subsubsection{Map \& Semantic Quality}
\label{sec:eval_ablation_oxford}
To evaluate the influence of the quality of the map and of the semantic segmentation used for localization, we conduct an ablation study on the Oxford dataset \cite{Maddern2017} using the training sequence from \cite{VonStumberg2019a}.
We choose three different voxel sizes for the map reconstruction to simulate different metric map qualities.
To evaluate different qualities of the semantic segmentation, we compare the results using the original network prediction (regular semantics) and a virtual rendering (improved semantics) generated from an accumulated 3D reconstruction of the tracking sequence while including dynamic objects from the original prediction.
The results for all configurations are presented in \refTable{tab:eval_ablation_oxford}.
It becomes apparent that higher quality maps denoted by lower voxel sizes consistently improve the localization performance.
Using the improved camera semantics also reduces the error, which can be explained by the fact that the original semantic prediction is not of particularly high quality considering the network was adapted to the target domain using only a couple of training images.

\begin{table}[th]
\caption{Median absolute translational error [cm] for different metric map and semantic quality on Oxford dataset \cite{Maddern2017}. Best value in each row marked as \textbf{bold}, best in each column marked as \underline{underlined}.}
\label{tab:eval_ablation_oxford}
\begin{center}
\begin{tabular}{lrrr}
\toprule
Map Voxel Size & 20cm & 10cm & 5cm \\
\midrule
Regular Semantics &  \underline{$0.407$} & $0.352$ & $\mathbf{0.328}$ \\
Improved Semantics & $0.439$ & \underline{$0.281$} & \underline{$\mathbf{0.208}$} \\
\bottomrule \end{tabular}\end{center}\end{table}\subsubsection{Semantic Map Content}
\begin{table}[th]
\caption{Median absolute translational error [cm] for different map elements. Best for each dataset denoted in \textbf{bold}. \xmark~ denotes tracking failure.}
\label{tab:eval_ablation_semantics}
\begin{center}
\setlength\tabcolsep{4.5pt}\begin{tabular}{lrrrrrrr}
\toprule
Road / Sidewalk & \cmark & \cmark & \cmark & \cmark & \cmark & \cmark & \cmark \\
Road Marking & \cmark & \cmark &        & \cmark & \cmark & \cmark & \cmark \\
Sign / Pole  & \cmark &        & \cmark & \cmark & \cmark & \cmark & \cmark \\
Barrier      &        & \cmark & \cmark & \cmark & \cmark & \cmark & \cmark \\
Building     &        &        &        &        & \cmark &        & \cmark \\
Nature       &        &        &        &        &        & \cmark & \cmark \\
\midrule
KITTI & \xmark & \xmark & \xmark & \xmark & $26.9$ & $18.4$ & $\mathbf{15.7}$ \\Oxford & $72.1$ & $99.9$ & $113.3$ & $52.0$ & $105.8$ & $45.2$ & $\mathbf{35.2}$ \\HD Map  & $\mathbf{20.6}$ & \xmark & \xmark & $22.3$ & $22.3$ & $23.1$ & $23.1$ \\\bottomrule \end{tabular}\end{center}\end{table}In addition to the map and semantic quality, different semantic object classes contained in the map may influence the performance. To evaluate the impact of different map elements on the localization performance, we dropped out individual classes from the maps. The results can be observed in \refTable{tab:eval_ablation_semantics}.
For the case of KITTI and Oxford we see that adding increasingly more map elements overall tends to improve results. Interestingly, for KITTI, the presence of seemingly unimportant \emph{building} or \emph{nature} classes is critical for proper tracking.
In additional experiments, we found that this is primarily because these map elements are important for determining the visibility of objects in the scene, i.e. a performance degradation can occur when occluded objects are rendered in the virtual view due to missing foreground objects in the map.
This is particularly important for the narrow suburban streets in KITTI.
On the other hand we observe in the case of the HD map that more map content does not always reduce the errors, which is most likely caused by inaccurately labeled map elements. We can further see that road markings, poles and signs play a crucial role, as tracking fails easily when these elements are not present.
Overall, we can conclude that more complete and accurate maps help to improve the localization accuracy and tracking robustness. \section{Conclusion}
In this paper, we have presented a novel approach for precise map-based vehicle localization by aligning semantically segmented images between camera frames and virtual map views for semantic consistency.
We have demonstrated that our proposed method is applicable to a wide range of different map formats from LiDAR-based mesh maps to generic high-definition maps, which do not need dedicated localization landmarks.
With median localization errors between 15 and 25cm in high-quality maps, our approach offers a localization accuracy sufficient for typical automated driving tasks, even in urban areas.
Our system performs particularly well when the environment is semantically rich and the map offers a high metric and semantic accuracy.
While the approach does not outperform recent works specifically tuned for image-based localization in terms of accuracy, it offers many other favorable properties for practical use, including real-time capability and low requirements for sensors and maps.
Considering that we only require semantically segmented input images with odometry and can reliably and accurately localize within a broad range of different map formats without special landmarks, we see great potential in our presented approach for automated vehicle localization.
Finally, further investigation with respect to robustness in less feature-rich rural or highway environments as well as robustness to domain shift in the semantics are an interesting avenue for future research.
 \section*{Acknowledgment}
We thank Patrick Wenzel for providing the evaluation and additional baseline results on the GN-Net Benchmark and all anonymous reviewers for their constructive feedback.

\bibliographystyle{bibliography/IEEEtran}
\bibliography{bibliography/IEEEabrv,export}

\end{document}